Original Paper

# Fine-tuned Sentiment Analysis of COVID-19 Vaccine–Related Social Media Data: Comparative Study

Chad A Melton[1,2], MSc; Brianna M White[2], MPH; Robert L Davis[2], MD, MPH; Robert A Bednarczyk[3], PhD; Arash Shaban-Nejad[2], MPH, PhD

[1]Bredesen Center for Interdisciplinary Research and Graduate Education, University of Tennessee at Knoxville, Knoxville, TN, United States
[2]Center for Biomedical Informatics, Department of Pediatrics, College of Medicine, University of Tennessee Health Science Center, Memphis, TN, United States
[3]Hubert Department of Global Health, Rollins School of Public Health, Emory University, Atlanta, GA, United States

**Corresponding Author:**
Arash Shaban-Nejad, MPH, PhD
Center for Biomedical Informatics
Department of Pediatrics, College of Medicine
University of Tennessee Health Science Center
50 N Dunlap Street, 492R
Memphis, TN, 38103
United States
Phone: 1 9012875836
Email: ashabann@uthsc.edu

## Abstract

**Background:** The emergence of the novel coronavirus (COVID-19) and the necessary separation of populations have led to an unprecedented number of new social media users seeking information related to the pandemic. Currently, with an estimated 4.5 billion users worldwide, social media data offer an opportunity for near real-time analysis of large bodies of text related to disease outbreaks and vaccination. These analyses can be used by officials to develop appropriate public health messaging, digital interventions, educational materials, and policies.

**Objective:** Our study investigated and compared public sentiment related to COVID-19 vaccines expressed on 2 popular social media platforms—Reddit and Twitter—harvested from January 1, 2020, to March 1, 2022.

**Methods:** To accomplish this task, we created a fine-tuned DistilRoBERTa model to predict the sentiments of approximately 9.5 million tweets and 70 thousand Reddit comments. To fine-tune our model, our team manually labeled the sentiment of 3600 tweets and then augmented our data set through back-translation. Text sentiment for each social media platform was then classified with our fine-tuned model using Python programming language and the Hugging Face sentiment analysis pipeline.

**Results:** Our results determined that the average sentiment expressed on Twitter was more negative (5,215,830/9,518,270, 54.8%) than positive, and the sentiment expressed on Reddit was more positive (42,316/67,962, 62.3%) than negative. Although the average sentiment was found to vary between these social media platforms, both platforms displayed similar behavior related to the sentiment shared at key vaccine-related developments during the pandemic.

**Conclusions:** Considering this similar trend in shared sentiment demonstrated across social media platforms, Twitter and Reddit continue to be valuable data sources that public health officials can use to strengthen vaccine confidence and combat misinformation. As the spread of misinformation poses a range of psychological and psychosocial risks (anxiety and fear, etc), there is an urgency in understanding the public perspective and attitude toward shared falsities. Comprehensive educational delivery systems tailored to a population's expressed sentiments that facilitate digital literacy, health information–seeking behavior, and precision health promotion could aid in clarifying such misinformation.

XSL·FO
RenderX

## Introduction

### Background

The novel coronavirus (COVID-19) has impacted and disrupted many aspects of everyday life worldwide. Following the implementation of rigid pandemic mitigation strategies in early 2020, social media use substantially increased with internet users turning to social media platforms to communicate and gather information regarding the dynamic and uncertain situation [1-4]. As the pandemic progressed and researchers worked to develop vaccines, many social media users turned their focus to gathering information regarding various topics related to COVID-19 vaccines, such as side effects, availability, and efficacy. As of May 19, 2022, approximately 6.27 million people across the world have died due to complications from COVID-19. Moreover, many experience long COVID syndrome, in which viral symptoms persist past the expected clinical recovery time [5]. Although COVID-19 vaccines are safe and effective at preventing life-threatening infections, hospitalizations, and deaths, vaccine hesitancy related to COVID-19 vaccines has led to further comorbidities and many preventable deaths [6-8].

With an estimated 4.5 billion users worldwide, social media offers an opportunity for near real-time analysis of large bodies of text data (500 million tweets/day) that could be useful to public health officials [3,9]. Using machine/deep learning, recent advancements in natural language processing methods (eg, Bidirectional Encoder Representations from Transformers [BERT], RoBERTa, GPT2, and XLNet) have substantially improved previous text classification models (greater than 90% accuracy) [4,10-14]. Moreover, pretrained models such as BERT or RoBERTa are available and free to researchers from platforms such as Hugging Face. These platforms are extremely helpful to the greater scientific community, considering that many of these models take several days on dozens of tensor processing units to learn [15,16]. Importantly, these models can be fine-tuned based on a particular use case (eg, text classification, text generation, and sentiment analysis). The enhanced functionality provides a researcher with techniques to investigate a wide variety of phenomena across many scientific domains [17-19]. Sentiment analysis (ie, classifying text as positive or negative) in particular is a powerful tool that can be used to correlate events to the public mood, surveil public health discussion, and even detect disease outbreaks [18]. Most importantly, these methods can be used by public health officials to develop precise messaging strategies and intervention campaigns to address the information crises and improve vaccination rates.

Our study sought to examine and explore sentiment regarding COVID-19 vaccines expressed on 2 popular social media platforms—Reddit and Twitter. We calculated positive and negative sentiment by creating a custom fine-tuned DistilRoBERTa model with data labeled by members of our team and then augmented by back-translation. We then offered a comparison of sentiment regarding COVID-19 vaccines across Reddit and Twitter. We hypothesized that we would observe somewhat similar trends in polarity between the 2 social media platforms with minor differences, because DistilRoBERTa has typically displayed accuracies greater than 90% [16]. However, we expected that our labeled data set would provide more nuanced insight into public sentiments in these 2 communities than previous sentiment analysis methods. Additionally, based on our previous work, we hypothesized that sentiment would remain more positive than negative [4]. Finally, we argued that identifying and following social media shared sentiment allows for the eventual development of comprehensive response strategies, which are better aimed at combatting misinformation and disinformation; improvement of vaccine delivery; and containment of disease transmission.

### COVID-19–Related Social Media Analysis

Social media content analysis is not a brand new concept and has been used for data mining and sentiment analysis before COVID-19. However, the nature of the pandemic response and the necessary separation of populations for safety have led to an unprecedented number of new users [9]. This influx caused a surge in social networking posts, leaving researchers with mountains of content to sort through. One positive aspect of social media data mining is that the content is publicly available and easily obtainable, allowing for rapid collection. The rapid collection of data, especially those related to COVID-19, permits researchers to follow the pandemic's progression alongside sentiment on the web. For example, the ability to rapidly collect tweets from a specified time period allows for the parallel analysis of general public opinion during major events, such as the release of the Pfizer vaccine in late 2020 or the death of a celebrity post–COVID-19 infection [20]. This targeted approach provides tools for niche discovery and exploration of the sentiment behind health decision-making.

Researchers have used the recent increase in opinion sharing to measure overall sentiment and vaccine hesitancy or acceptance [4,20-24]. As social media usage has continued to grow throughout the pandemic era, more than 3.6 billion people are known to regularly log on to at least one networking platform. Twitter is considered one of the largest and most used social media platforms, with more than 400 million account owners [9]. The platform allows users to post short messages or tweets for "followers" to see and respond to, based on the underlying sentiment they evoke. Tweets are limited to brief messages, with a 280-character limit, but may contain attached images, videos, or highlighted popular keywords known as "hashtags." Additionally, tweets can include hyperlinks to news articles or scientific literature. If another user agrees with a posted tweet, they can "retweet" or share the message to their profiles in a show of rapport. Rather than joining topic-based communities, users typically follow other users.

The Reddit platform is similar in size, with approximately 430 million current users [9]. However, it is different in message format and delivery, in that users are allowed to create groupings based on a topic, called "subreddits." Subreddits often contain open dialogue alongside images, videos, and hyperlinks to news articles or literature. Similar to "retweeting," subreddit subscribers have the unique ability to "upvote" or "downvote" a post based on the user's opinion of its contents. Users are also able to join the discussion by leaving comments, which can also be upvoted or downvoted. If a subreddit becomes increasingly popular and receives a good share of upvotes, the post will appear first within a topic category. The more traffic a subreddit receives, even if it is sharing misinformation or disinformation, the higher the Reddit platform will promote it. Notably, subreddits generally have rules that community members must adhere to or risk the potential for the removal of a post or banning.

### BERT Algorithm

Substantial advances in natural language processing have occurred since the development of BERT and the work built from its architecture. BERT is a powerful and versatile artificial intelligence–based natural language processing algorithm developed at Google AI Language that excels at text classification (ie, ontologies, categories, and sentiment, etc) of unstructured/semistructured text data that are characteristic of social media data [10]. The BERT algorithm was trained on the entirety of Wikipedia and the Brown Corpus over 4 days using 16 cloud-based tensor processing units. BERT is a transformer-based language model that uses multiple encoders to create word embeddings. These embeddings are then used in concert with masked language modeling and next sentence prediction to learn by predicting random masked words in a sentence and learning to predict sentences, respectively. These 2 steps teach BERT to understand context, a skill that older recurrent neural networks typically struggled with. A convenient aspect of BERT is that it has the capability to fine-tune the model with relevant data by replacing the output layer with weights from custom data. Researchers have been inspired by the original BERT architecture to create many variations (eg, RoBERTa, DistilRoBERTa, DistilBERT, and BART, etc) that have surpassed the benchmarks of previous models. Moreover, these models can be fine-tuned for specific domain-based tasks (ClinicalBERT and BioBERT) in multiple languages [11,12,25]. Furthermore, several studies have used other fine-tuned BERT models to investigate COVID-19–related content expressed on social media related to misinformation detection, sentiment classification, and continent analysis [13,26-29].

## Methods

### Study Overview

Our study compared COVID-19 vaccine–related postings from 2 popular social media platforms—Reddit and Twitter—from January 1, 2020, to March 1, 2022. These 2 platforms were chosen due to their worldwide usage, vibrant discussions, and high user count. The time frame included the earliest parts of the pandemic to trace the evolution of sentiments over time. Most importantly, these platforms were chosen because only a small number of comparative studies have focused on the typical user, especially studies related to COVID-19 vaccine sentiment or other vaccines. Our study used a binary (ie, positive or negative polarity) sentiment classification method for training our model and for sentiment analysis. A binary system was chosen for a few reasons. (1) Binary systems are more computationally efficient when processing large bodies of data. (2) Binary classifiers are typically more accurate than multiclass systems. (3) In the past, sentiment classifiers that incorporate a neutral class often rely on a low probability or confidence score. Since our model reported a confidence value, this information could be extrapolated.

### Data Overview

Substantial effort was taken to identify and remove Twitter posts that were found to be directly from news agencies or bots. These posts were identified by their source having an overwhelmingly high post count during the 26-month period relative to the average number of posts of a "normal" user, as well as by visually inspecting tweets of users that appeared at an abnormal frequency. Both Twitter and Reddit data sets were limited to only include users who posted fewer than or equal to 200 times throughout our time frame. These steps were important due to the repetitive nature of many bot tweets, which had the potential to skew sentiment calculations and misalign the goal to compare the normal user base of both platforms. Although the methodologies in harvesting Reddit and Twitter data differ slightly, both data sets underwent similar cleaning steps. Both data sets were queried for the same relevant terms typically present in web-based discussions about COVID-19 vaccines. This step was important due to the tendency for some extended comment threads to meander off-topic. This occurrence was especially true with threads from some Reddit communities. The daily posting frequencies of the 2 platforms were relatively similar in the early months of the pandemic. The frequency increased dramatically for both platforms in late September to October 2020 as news of vaccine circulation became more widespread. Although each platform displayed 4 spikes in posting frequency at similar time periods (October 2020, March to April 2021, August to September 2021, and December 2021 to January 2022), they obtained a maximum in different time periods. Reddit reached its maximum posting from March to April 2021, whereas Twitter reached its maximum from September to October 2021.

### Twitter

Approximately 13 million tweets were harvested using the *snscrape* and *Tweepy* API Python libraries based on the search term "COVID Vaccine." After removing tweets by suspected bots, news media, or highly repetitive high-frequency users and duplicate tweets, our final Twitter data set consisted of 9,518,270 tweets authored by 3,006,075 Twitter users. The tweets contained approximately 16.32 million total likes, with a maximum of 430,758 likes and an average of 14.9 likes per tweet. Tweets cannot be downvoted, but approximately 4,794,865 tweets were attributed with 0 likes. Statistics on tweet sharing or retweets were not collected because this metric was not available for both platforms.

### Reddit

We harvested 579,241 user-created posts from 67 subreddits with the Python Reddit API *Wrapper*. These subreddits were collected to gain a broad understanding of sentiments related to the COVID-19 vaccines as well as to avoid potential biases in data collection. These subreddits contained a total of 5,590,913 subscribers as of March 1, 2022. Our query removed a large portion of unrelated terms. After visually inspecting and confirming the results of the querying process, our final Reddit data set consisted of 67,962 comments composed by at least 9843 authors. These posts contained approximately 2.1 million total upvotes, with an average of 31 upvotes and a maximum of 18,253 upvotes per comment.

### Data Labeling and Augmentation

Since time is of the essence in a global pandemic, combined with the fact that labeling data is time-consuming and costly, we created a custom training data set by labeling sentiment (positive or negative) for approximately 3600 tweets related to COVID-19 vaccines. We chose to label tweets exclusively for this study, because the 280-character limit of a tweet (ie, compared to a Reddit post limit of a maximum of 10,000 characters) would allow our small team to create a time-relevant training data set more quickly. We then augmented our data set through the process of back-translation with several language models on the Hugging Face model repository. Back-translation was chosen after testing a few other methods of text augmentation. Some techniques (eg, word masking) resulted in far more duplicated texts that would eventually need to be removed. Back-translation relies on subtle differences between language structure, word meaning, and syntax. In effect, the outputted text will vary slightly from the inputted text without losing semantic and contextual meaning [14]. In our case, the back-translation method translated our English-language text into another language (eg, French, Chinese, Greek, and Hebrew) and then back into English. After removing duplicates, our final augmented data set consisted of 48,691 tweets.

### RoBERTa and DistilRoBERTa

For our study, we chose to explore the capabilities of DistilRoBERTa. RoBERTa is a more robust model than BERT, and DistilRoBERTa is an optimized version of RoBERTa [15,16]. Developed at Facebook, RoBERTa was trained on 160 GB of text compared to the 16 GB of BERT. RoBERTa dropped the next sentence prediction feature of BERT and added dynamic token masking during training. These enhancements are estimated to have improved the original BERT's performance significantly (2% to 20%) [16]. Compared to RoBERTa, DistilRoBERTa was trained on approximately 40 GB of text data (OpenWebTextCorpus) and operates about twice as fast.

### The University of Tennessee Health Science Center Vaccine Sentiment Labeling and DistilRoBERTa Fine-tuning

We fine-tuned the DistilRoBERTa base via the Hugging Face *Trainer* class, which provides the user with an API for training with *PyTorch*. Our data were then randomized and segregated into 40,000 training tweets, 4000 validation tweets, and 4691 tweets for testing. Training hyperparameters included a $2 \times 10^{-5}$ learning rate, 32 training and evaluation batch size, a seed number of 42, and a linear scheduler with 500 warm-up steps. We used the *Adam* optimizer with betas of 0.9 and 0.999 and an epsilon of $1 \times 10^{-8}$. Lastly, our model was trained for 2 epochs. These hyperparameters achieved a training loss of 0.1284, a validation loss of 0.1167, a precision of 0.9561, an $F_1$-score of 0.9592, and an accuracy of 0.9592 (see Table 1).

Table 1. DistilRoBERTa fine-tuning training metrics. The model obtained optimal fine-tuning after 2 training epochs.

| Step | Epoch | Training loss | Validation loss | Precision | Accuracy | $F_1$-score |
| --- | --- | --- | --- | --- | --- | --- |
| 500 | 0.4 | 0.5903 | 0.4695 | 0.7342 | 0.7728 | 0.7890 |
| 1000 | 0.8 | 0.3986 | 0.3469 | 0.8144 | 0.8596 | 0.8684 |
| 1500 | 1.2 | 0.2366 | 0.1939 | 0.9313 | 0.9260 | 0.9253 |
| 2000 | 1.6 | 0.1476 | 0.1560 | 0.9207 | 0.9452 | 0.9465 |
| 2500 | 2.0 | 0.1284 | 0.1167 | 0.9561 | 0.9592 | 0.9592 |

### Analytical Methods

Following the fine-tuning of our model, we processed the Twitter and Reddit data through the Hugging Face *pipeline* for sentiment analysis. The model returned a label of either positive or negative for each tweet or Reddit comment. Along with the determined polarity, the model also returned a probabilistic confidence score ranging from 0 to 1. For clarity, tweets or comments classified as negative were multiplied by –1 to reflect the negative sentiment.

### Ethical Considerations

No ethical approval was needed from our institution due to the public availability and nonidentifiable nature of the data used.

## Results

### DistilRoBERTa Fine-tuned to COVID-19 Vaccine

*Twitter*

The DistilRoBERTa fine-tuned polarity analysis determined that the 9,518,270 tweets were more negative (n=5,215,830, 54.8%) than positive (n=4,302,440, 45.2%) throughout our time frame (see Figure 1).

The maximum positive rating occurred in March 2021 (375,789/675,274 55.6%). However, the minimum positive rating occurred in January 2022 (191,159/526,582, 36.3%), displaying a steady decrease in polarity from the maximum. For the confidence score, the tweets classified as positive had a maximum score of 0.999, a minimum of approximately 0 ($3.58 \times 10^{-7}$), and a mean of 0.868 (see Figure 2). The tweets classified as negative had a minimum score of –0.999, a maximum value of approximately zero ($-1.78 \times 10^{-6}$), and a mean of –0.882 (see Figures 1 and 2).

**Figure 1.** Tweet polarity from the DistilRoBERTa model fine-tuned to COVID-19 vaccine. Polarity and the corresponding confidence probability are represented on the y-axis, and time is represented on the x-axis. Tweets are represented as light blue circles. Circle size indicates the number of likes per tweet—larger circles indicate more likes and smaller circles indicate fewer likes.

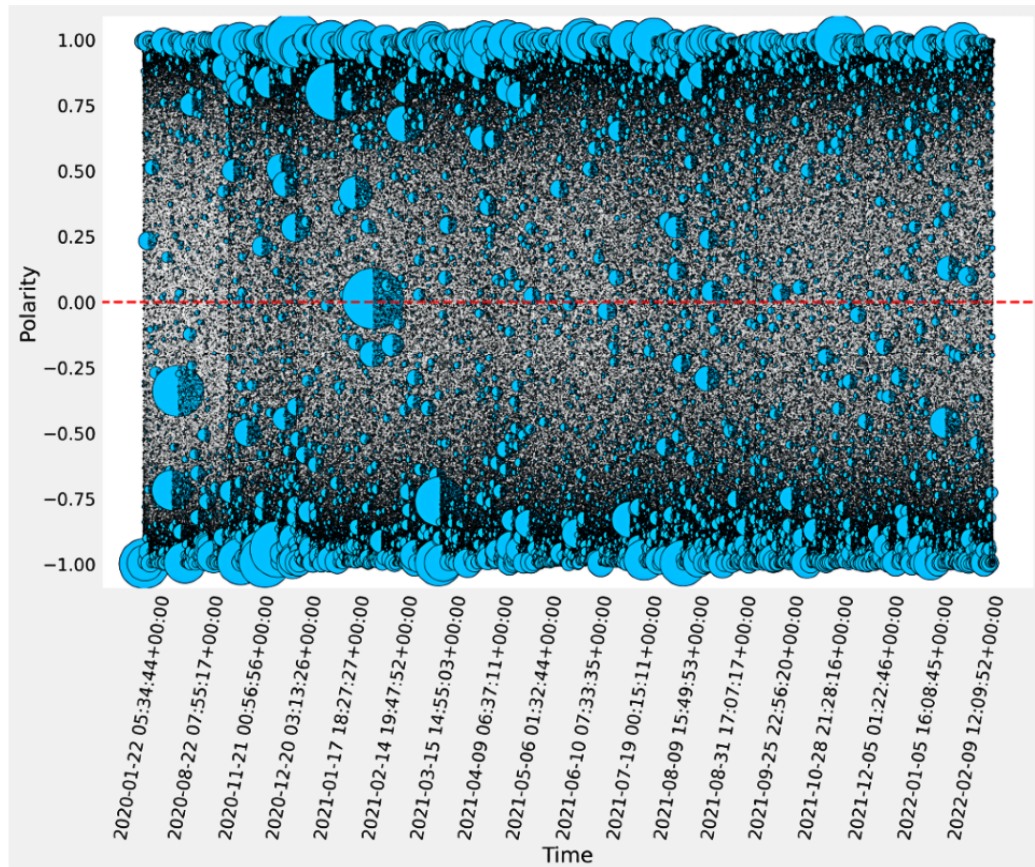

**Figure 2.** Confidence score versus like count for Twitter. The x-axis represents the confidence score and the y-axis represents the number of likes a tweet received. Data points below 0.00 on the x-axis represent a negative classification, and data points above 0.00 represent a positive classification. Data points are represented as light blue circles.

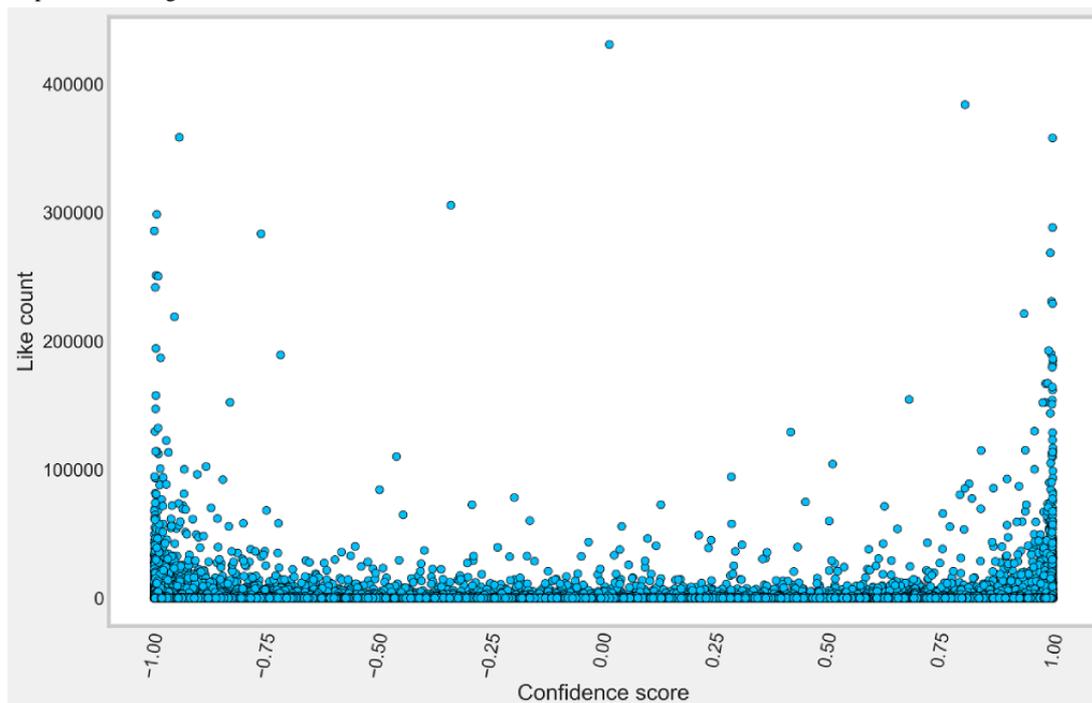

### Reddit

The Reddit sentiment polarity analysis for the fine-tuned DistilRoBERTa model found that of the 67,962 posts, 37.7% (n=25,646) were classified as negative and 62.3% (n=42,316) were classified as positive. The highest polarity reported in our experiment and the maximum positive rating occurred in April 2021 (6611/9044, 73.1 %), and the minimum positive rating occurred in February 2020 (170/351, 48.4%). For the confidence scores, the comments classified as positive had a maximum score of 0.999, a minimum of approximately 0 ($1.55 \times 10^{-4}$), and a mean of 0.870 (see Figure 3). The comments classified as negative had a minimum of –0.999, a maximum of approximately 0 ($-4.74 \times 10^{-5}$), and a mean of –0.808 (see Figures 3 and 4).

**Figure 3.** Reddit comment polarity from the DistilRoBERTa model fine-tuned to COVID-19 vaccine. Polarity and corresponding confidence probability are represented on the y-axis, and time is represented on the x-axis. Data points are represented as orange-red circles. Circle size indicates the number of upvotes per comment—more upvotes are represented by larger circles and fewer upvotes are represented by smaller circles.

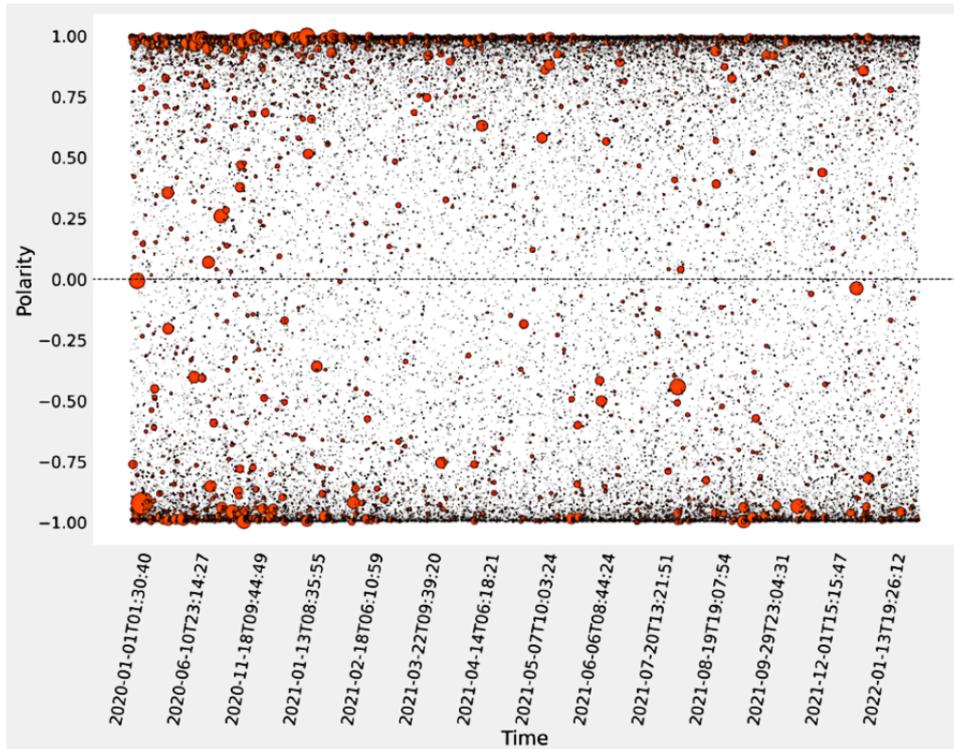

**Figure 4.** Confidence score versus like count for Reddit. The x-axis represents the confidence score and the y-axis represents the number of upvotes a comment received. Data points below 0.00 on the x-axis represent a negative classification, and data points above 0.00 represent a positive classification. Data points are represented as orange-red circles.

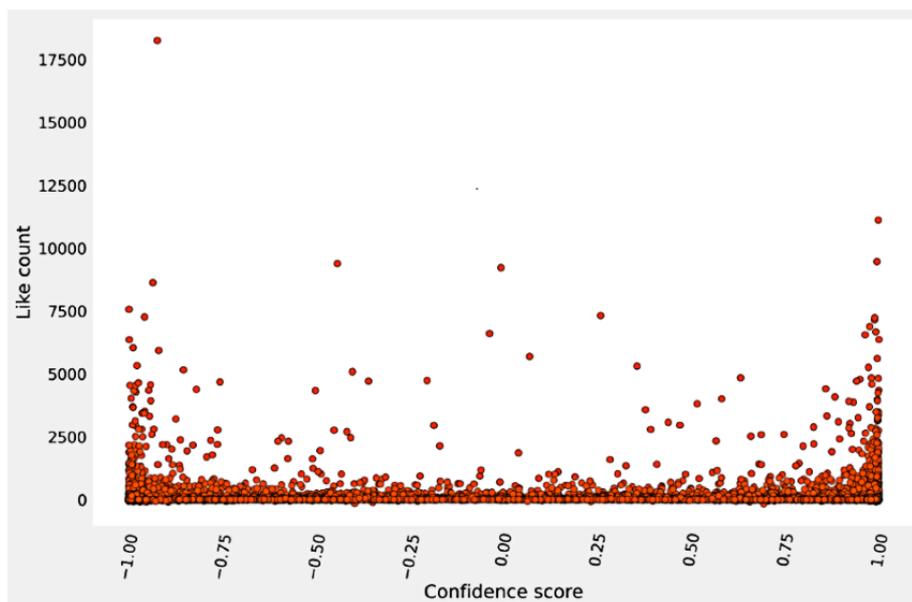

## COVID-19 Vaccine Sentiment Expressed on Reddit and Twitter

Overall, the average sentiment for the 2 social media platforms was somewhat different (62.3% positive on Reddit vs 45.2% positive on Twitter). An interesting story begins to appear when looking closely at the month-to-month results in relation to each other. Although sentiment on both platforms oscillated in the early months of the pandemic, Reddit sentiment was higher (ranging from 48% to 55% positive) from January to August 2020. Twitter sentiment began similar to Reddit sentiment but gradually declined until becoming substantially more negative from September to October 2020, and then increasing to a maximum of 55% in March 2021. Reddit sentiment began a steep increase in polarity in December 2020 and continued to increase until reaching the maximum positive sentiment (approximately 73%) in April 2021. After sentiment on each platform achieved their maximum positive polarity, both began an oscillating and gradual decline in sentiment to near early pandemic levels. However, Twitter sentiment continued to fall until achieving a minimum of 36% (see Figure 5).

**Figure 5.** Monthly sentiment for Twitter and Reddit COVID-19 vaccine–related posts. The x-axis represents time and the y-axis represents the percentage of posts classified as positive. The blue line represents Twitter sentiment and the orange-red line represents Reddit sentiment. Note that since posting frequency was very low, sentiment for January 2020 is an average of all other months of corresponding data.

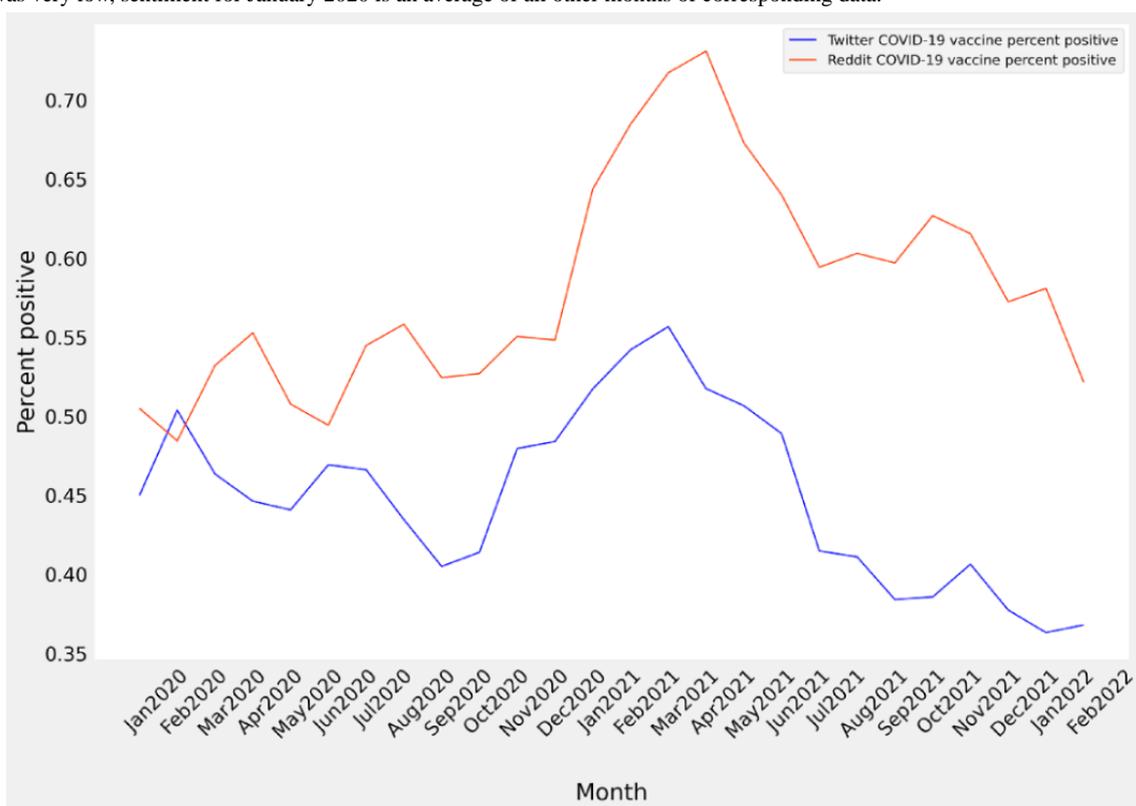

## Discussion

### Interpretation of Results

Ranging from January 1, 2020, to March 1, 2022, our results show that the average sentiment for the Reddit data set was more positive than the average sentiment expressed on Twitter. Interestingly, both platforms expressed similar sentiment changes during key moments of the pandemic (eg, vaccine efficacy announcements, vaccine distribution to all ages, new variants, and waning efficacy). This behavior is especially observable as vaccines became widely available to the public and the polarity diminished. Considering this similar behavior, we feel that both Twitter and Reddit continue to be valuable data sources that public health officials can use to develop vaccine education campaigns and digital interventions. Although Twitter is superior in the ability to access large numbers of tweets through an API, substantial steps need to be taken while cleaning Twitter data to remove bots, news media posts, commercial users, duplicates, and users who have extremely high posting frequencies. On the other hand, Reddit data are more plentiful in longer texts that could be more useful for topic modeling.

What drove sentiment changes related to COVID-19 vaccines on these 2 platforms? One possibility could be related to the character limit of tweets versus Reddit posts (ie, 280 vs 10,000 characters, respectively). The shortened character limit of tweets most likely contributes to the quick spread of information and can be reactionary in nature, driving negative sentiment. However, Reddit users typically take advantage of the longer character limit and share, at times, highly personal stories and experiences related to their health care. For this reason, Reddit could remain a highly valuable source when considering the development of public health messaging and education campaigns.

Correlating changes in sentiment with developments during the pandemic presents some interesting challenges and ideas alike. The most obvious steep increase in sentiment seems to be correlated with positive news regarding vaccine development and trials and news of high efficacy, distribution, and availability to those who patiently waited for the vaccine. It is challenging to correlate minimum sentiment scores because their decline was not uniform. It is highly likely that the gradual decline was related to a combination of unfortunate events related to the pandemic (eg, misinformation, pandemic fatigue, and falling vaccine efficacy). It is conceivable that challenges in vaccine rollout and distribution could negatively affect sentiment. However, previous topic modeling and semantic network analysis on portions of this data set did not find a meaningful occurrence of terms related to vaccine distribution. Therefore, more psychological, sociological, and cultural studies are desperately needed to understand what drives certain populations, news media, politicians, and entertainers to so readily accept and propagate misinformation and conspiracy theories rather than directly observable facts. Such studies would not only benefit future public health responses but also many other areas of life where misinformation and disinformation have taken hold. The success of digital interventions and education campaigns would likely be limited without a more thorough understanding of how to reach these populations.

## Public Health Implications

The application of our findings could have momentous impacts on the public health sector in the fight against infectious diseases such as COVID-19. Further development of low–human effort surveillance systems optimized for the rapid collection of data would allow for the real-time analysis of public emotion in correlation with disease progression. Moreover, fine-tuning models to assess geographical and demographical differences in sentiment could provide insight into the attitudes of populations at the greatest risk of debilitating outcomes. In addition to geographically and demographically specific data mining, targeting public discourse during times of peak infection, vaccine releases, or the death of a celebrity, athlete, or political figure due to the disease could greatly bolster public health response [30,31]. The expansion of such disease projection and prediction models using sentiment mining techniques could also influence evidence-informed policy. Discerning the dynamic levels of population sentiment allows public health officials to design catered policy communication strategies. By providing the necessary tools to better understand public emotion related to disease prevention, control, and containment, policy makers would be better equipped to evaluate program successes and highlight any need for repositioning.

Furthermore, the analysis of sentiment shared via social media could prove to be a vital instrument in combatting rampant misinformation and disinformation shared on the web. As the spread of misinformation poses a range of psychological and psychosocial risks (anxiety and fear, etc), there is an urgency in understanding the public perspective and attitude toward shared falsities. Education delivery systems tailored to population-expressed sentiment could aid in clarifying such misinformation. Moreover, there is room for the expansion of artificially intelligent messaging systems, tasked with generating responses to waves of misinformation and disinformation shared via social media platforms. Overall, the proposed framework for the real-time analysis of sentiment could be useful in guiding governmental support of public health recovery efforts.

## Limitations

As with most studies, ours has some limitations. Challenges occur when conducting sentiment analysis in social media texts due to some long-standing problems. Although BERT and newer models greatly mitigate many of these challenges, some models typically struggle with detecting sarcasm, humor, emotion, and complex inferences in texts unless specifically having been trained to do so. For example, many pro-vaccine social media users express extremely negative views and sentiments regarding the anti-vaccine community. How would BERT classify such an occurrence? Although their expressed sentiment is positive toward the vaccine, many natural language processing algorithms and data labelers would potentially struggle with this type of classification. Even though we took great care with this study to remove tweets by bots or tweets from highly repetitive users from Twitter and choose unbiased subreddits, it is possible that some could have still slipped through the data cleaning process. Moreover, augmented data can potentially cause problems with overfitting when fine-tuning models due to relatively similar semantic content. We limited our training epochs and closely monitored the relationship between training loss and validation loss to mitigate this potential problem. Future work could involve efforts to create a larger labeled data set that would include not only COVID-19 vaccine sentiments but those of other vaccines as well.

## Conclusions

We conducted a sentiment analysis of approximately 70,000 Reddit comments and 9.5 million tweets with a fine-tuned DistilRoBERTa model. Our analysis found that both Reddit and Twitter users expressed similar changes in sentiment throughout the pandemic, even though Twitter was substantially more negative than Reddit. Although subtle differences in sentiment were observed monthly, both platforms demonstrated a substantial increase in positive sentiments as the COVID-19 vaccine became readily available to the general public. The results we present here are a portion of an ongoing study to investigate vaccine-related content on social media with a focus on identifying and combating misinformation in efforts to decrease vaccine hesitancy. Correlating strong sentiment with high infectivity rates could provide officials with forecasting for the public acceptance of migration strategies such as vaccine delivery and uptake. These integrated disease surveillance tools should not only be leveraged in the fight against COVID-19 but stand to play essential roles in the evolution of future health policy, decision-making, program implementation, and precision health promotion [32]. In the near future, our team plans to expand the methods demonstrated in this study into sentiment related to other types of vaccines (eg, human papillomavirus vaccines). We expect these results along with others to be used to develop tools to assist public health officials in monitoring public discourse regarding disease outbreaks, gaining a better understanding of vaccine hesitancy, and developing personalized digital interventions [33,34] and education campaigns.

## Acknowledgments

We would like to thank our team of data labelers from the University of Tennessee Health Science Center. This study is partially supported by a grant (1R37CA234119-01A1) from the National Cancer Institute (NCI).

## Data Availability

The data that support our findings are available upon reasonable request to the authors. Data are not available for commercial use.

## Authors' Contributions

CAM conceptualized and supervised the study and drafted, reviewed, and edited the manuscript. BMW conceptualized the study and drafted, reviewed, and edited the manuscript. RLD reviewed and edited the manuscript. RAB reviewed and edited the manuscript. ASN drafted, reviewed, and edited the manuscript; supervised the study; and acquired funding.

## Conflicts of Interest

None declared.

## References

1. Sharma C, Whittle S, Haghighi PD, Burstein F, Keen H. Sentiment analysis of social media posts on pharmacotherapy: a scoping review. Pharmacol Res Perspect 2020 Oct 19;8(5):e00640 [FREE Full text] [doi: 10.1002/prp2.640] [Medline: 32813329]
2. Bento AI, Nguyen T, Wing C, Lozano-Rojas F, Ahn Y, Simon K. Evidence from internet search data shows information-seeking responses to news of local COVID-19 cases. Proc Natl Acad Sci U S A 2020 May 26;117(21):11220-11222 [FREE Full text] [doi: 10.1073/pnas.2005335117] [Medline: 32366658]
3. Auxier B, Anderson M. Social media use in 2021. Pew Research Center. 2021 Apr 07. URL: https://www.pewresearch.org/internet/2021/04/07/social-media-use-in-2021/ [accessed 2022-03-20]
4. Melton CA, Olusanya OA, Ammar N, Shaban-Nejad A. Public sentiment analysis and topic modeling regarding COVID-19 vaccines on the Reddit social media platform: a call to action for strengthening vaccine confidence. J Infect Public Health 2021 Oct;14(10):1505-1512 [FREE Full text] [doi: 10.1016/j.jiph.2021.08.010] [Medline: 34426095]
5. Raveendran AV, Jayadevan R, Sashidharan S. Long COVID: an overview. Diabetes Metab Syndr 2021 May;15(3):869-875 [FREE Full text] [doi: 10.1016/j.dsx.2021.04.007] [Medline: 33892403]
6. Rosenberg ES, Holtgrave DR, Dorabawila V, Conroy M, Greene D, Lutterloh E, et al. New COVID-19 cases and hospitalizations among adults, by vaccination status - New York, May 3-July 25, 2021. MMWR Morb Mortal Wkly Rep 2021 Sep 17;70(37):1306-1311 [FREE Full text] [doi: 10.15585/mmwr.mm7037a7] [Medline: 34529645]
7. WHO coronavirus (COVID-19) dashboard. World Health Organization. URL: https://covid19.who.int/ [accessed 2022-03-20]
8. Benham JL, Atabati O, Oxoby RJ, Mourali M, Shaffer B, Sheikh H, et al. COVID-19 vaccine-related attitudes and beliefs in Canada: national cross-sectional survey and cluster analysis. JMIR Public Health Surveill 2021 Dec 23;7(12):e30424 [FREE Full text] [doi: 10.2196/30424] [Medline: 34779784]
9. Clement J. Most popular social networks worldwide as of January 2022, ranked by number of monthly active users. Statista. 2022 Jan. URL: https://www.statista.com/statistics/272014/global-social-networks-ranked-by-number-of-users/ [accessed 2022-03-20]
10. Devlin J, Chang MW, Lee K, Toutanova K. Bert: pre-training of deep bidirectional transformers for language understanding. 2019 Jun Presented at: 2019 Conference of the North American Chapter of the Association for Computational Linguistics: Human Language Technologies; June 2-7, 2019; Minneapolis, MN p. 4171-4186. [doi: 10.18653/v1/N19-1423]
11. Alsentzer E, Murphy J, Boag WH, Weng W, Jin D, Naumann T, et al. Publicly available clinical BERT embeddings. 2019 Jun Presented at: 2nd Clinical Natural Language Processing Workshop; June 7, 2019; Minneapolis, MN p. 72-78. [doi: 10.18653/v1/w19-1909]
12. Lee J, Yoon W, Kim S, Kim D, Kim S, So CH, et al. BioBERT: a pre-trained biomedical language representation model for biomedical text mining. Bioinformatics 2020 Feb 15;36(4):1234-1240 [FREE Full text] [doi: 10.1093/bioinformatics/btz682] [Medline: 31501885]
13. Kolluri N, Liu Y, Murthy D. COVID-19 misinformation detection: machine-learned solutions to the infodemic. JMIR Infodemiology 2022 Aug 25;2(2):e38756. [doi: 10.2196/38756]
14. Beddiar DR, Jahan MS, Oussalah M. Data expansion using back translation and paraphrasing for hate speech detection. Online Soc Netw Media 2021 Jul;24:100153. [doi: 10.1016/j.osnem.2021.100153]
15. Liu Y, Ott M, Goyal N, Du J, Joshi M, Chen D, et al. Roberta: a robustly optimized bert pretraining approach. ArXiv 2019 Jul 26:1-13. [doi: 10.48550/arXiv.1907.11692]
16. Sanh V, Debut L, Chaumond J, Wolf T. DistilBERT, a distilled version of BERT: smaller, faster, cheaper and lighter. ArXiv 2019 Oct 02:1-5. [doi: 10.48550/arXiv.1910.01108]

17. Blane JT, Bellutta D, Carley KM. Social-cyber maneuvers during the COVID-19 vaccine initial rollout: content analysis of tweets. J Med Internet Res 2022 Mar 07;24(3):e34040 [FREE Full text] [doi: 10.2196/34040] [Medline: 35044302]
18. Singh R, Singh R, Bhatia A. Sentiment analysis using machine learning technique to predict outbreaks and epidemics. Int J Adv Sci Res 2018 Mar;3(2):19-24 [FREE Full text]
19. Tang L, Bie B, Park S, Zhi D. Social media and outbreaks of emerging infectious diseases: a systematic review of literature. Am J Infect Control 2018 Sep;46(9):962-972 [FREE Full text] [doi: 10.1016/j.ajic.2018.02.010] [Medline: 29628293]
20. Jennings W, Stoker G, Bunting H, Valgarðsson VO, Gaskell J, Devine D, et al. Lack of trust, conspiracy beliefs, and social media use predict COVID-19 vaccine hesitancy. Vaccines (Basel) 2021 Jun 03;9(6):593 [FREE Full text] [doi: 10.3390/vaccines9060593] [Medline: 34204971]
21. Puri N, Coomes EA, Haghbayan H, Gunaratne K. Social media and vaccine hesitancy: new updates for the era of COVID-19 and globalized infectious diseases. Hum Vaccin Immunother 2020 Nov 01;16(11):2586-2593 [FREE Full text] [doi: 10.1080/21645515.2020.1780846] [Medline: 32693678]
22. Shakir A, Arora J. Accuracy in binary, ternary and multi-class classification sentimental analysis-a survey. Int J Advanced Res Comput Sci 2018 Feb 20;9(2):524-526. [doi: 10.26483/ijarcs.v9i2.5866]
23. Hou Z, Tong Y, Du F, Lu L, Zhao S, Yu K, et al. Assessing COVID-19 vaccine hesitancy, confidence, and public engagement: a global social listening study. J Med Internet Res 2021 Jun 11;23(6):e27632 [FREE Full text] [doi: 10.2196/27632] [Medline: 34061757]
24. Kummervold PE, Martin S, Dada S, Kilich E, Denny C, Paterson P, et al. Categorizing vaccine confidence with a transformer-based machine learning model: analysis of nuances of vaccine sentiment in Twitter discourse. JMIR Med Inform 2021 Oct 08;9(10):e29584 [FREE Full text] [doi: 10.2196/29584] [Medline: 34623312]
25. Yum Y, Lee JM, Jang MJ, Kim Y, Kim J, Kim S, et al. A word pair dataset for semantic similarity and relatedness in Korean medical vocabulary: reference development and validation. JMIR Med Inform 2021 Jun 24;9(6):e29667 [FREE Full text] [doi: 10.2196/29667] [Medline: 34185005]
26. Ayoub J, Yang XJ, Zhou F. Combat COVID-19 infodemic using explainable natural language processing models. Inf Process Manag 2021 Jul;58(4):102569 [FREE Full text] [doi: 10.1016/j.ipm.2021.102569] [Medline: 33776192]
27. Koren A, Alam MAU, Koneru S, DeVito A, Abdallah L, Liu B. Nursing perspectives on the impacts of COVID-19: social media content analysis. JMIR Form Res 2021 Dec 10;5(12):e31358 [FREE Full text] [doi: 10.2196/31358] [Medline: 34623957]
28. Hussain A, Tahir A, Hussain Z, Sheikh Z, Gogate M, Dashtipour K, et al. Artificial intelligence-enabled analysis of public attitudes on Facebook and Twitter toward COVID-19 vaccines in the United Kingdom and the United States: observational study. J Med Internet Res 2021 Apr 05;23(4):e26627 [FREE Full text] [doi: 10.2196/26627] [Medline: 33724919]
29. Cresswell K, Tahir A, Sheikh Z, Hussain Z, Domínguez Hernández A, Harrison E, et al. Understanding public perceptions of COVID-19 contact tracing apps: artificial intelligence-enabled social media analysis. J Med Internet Res 2021 May 17;23(5):e26618 [FREE Full text] [doi: 10.2196/26618] [Medline: 33939622]
30. Salali GD, Uysal MS. Effective incentives for increasing COVID-19 vaccine uptake. Psychol Med 2021 Sep 20:1-3 [FREE Full text] [doi: 10.1017/S0033291721004013] [Medline: 34538287]
31. Romaniuc R, Guido A, Mai N, Spiegelman E, Sutan A. Increasing vaccine acceptance and uptake: a review of the evidence. SSRN Preprint posted online on May 11, 2021. [doi: 10.2139/ssrn.3839654]
32. Shaban-Nejad A, Michalowski M, Peek N, Brownstein JS, Buckeridge DL. Seven pillars of precision digital health and medicine. Artif Intell Med 2020 Mar;103:101793. [doi: 10.1016/j.artmed.2020.101793] [Medline: 32143798]
33. Olusanya OA, Ammar N, Davis RL, Bednarczyk RA, Shaban-Nejad A. A digital personal health library for enabling precision health promotion to prevent human papilloma virus-associated cancers. Front Digit Health 2021 Jul 21;3:683161 [FREE Full text] [doi: 10.3389/fdgth.2021.683161] [Medline: 34713154]
34. Olusanya OA, White B, Melton CA, Shaban-Nejad A. Examining the implementation of digital health to strengthen the COVID-19 pandemic response and recovery and scale up equitable vaccine access in African countries. JMIR Form Res 2022 May 17;6(5):e34363 [FREE Full text] [doi: 10.2196/34363] [Medline: 35512271]

## Abbreviations

**BERT:** Bidirectional Encoder Representations from Transformers